\documentclass[11pt,a4paper]{article}
\usepackage[hyperref]{naaclhlt2019}
\usepackage{times}
\usepackage{latexsym}
\usepackage{graphicx}
\usepackage{xspace}
\usepackage{url}
\usepackage{subcaption}
\usepackage{pgfplots}
\usepackage{amsmath}
\usepackage{cleveref}

\DeclareMathOperator{\sigmoid}{\sigma}

\newcommand{\ndcg}{\ensuremath{\textit{nDCG}}\xspace}

\newcommand{\bigruatt}{\textsc{bigru-att}\xspace}

\newcommand{\bigru}{\textsc{bigru}\xspace}
\newcommand{\gru}{\textsc{gru}\xspace}

\newcommand{\rnn}{\textsc{rnn}\xspace}
\newcommand{\cnn}{\textsc{cnn}\xspace}
\newcommand{\han}{\textsc{han}\xspace}

\newcommand{\maxpool}{\textsc{maxpool}\xspace}
\newcommand{\lstm}{\textsc{lstm}\xspace}

\newcommand{\xmtc}{\textsc{xmtc}\xspace}

\newcommand{\lwan}{\textsc{lwan}\xspace}

\newcommand{\lwancnn}{\textsc{cnn-lwan}\xspace}
\newcommand{\zlwancnn}{\textsc{z-cnn-lwan}\xspace}
\newcommand{\lwangru}{\textsc{bigru-lwan}\xspace}

\newcommand{\zlwangru}{\textsc{z-bigru-lwan}\xspace}

\newcommand{\maxhss}{\textsc{max-hss}\xspace}
\newcommand{\lwhan}{\textsc{lw-han}\xspace}
\newcommand{\glove}{\textsc{glove}\xspace}
\newcommand{\ensemble}{\textsc{ensemble-lwan}\xspace}
\newcommand{\matchcnn}{\textsc{match-cnn}\xspace}
\newcommand{\newdata}{\textsc{eurlex57k}\xspace}

\newcommand{\eurovoc}{\textsc{eurovoc}\xspace}
\newcommand{\eu}{\textsc{eu}\xspace}
\newcommand{\rcv}{\textsc{rcv1}\xspace}
\newcommand{\icd}{\textsc{icd-9}\xspace}
\newcommand{\eurlex}{\textsc{eur-lex}\xspace}
\newcommand{\cellar}{\textsc{cellar}\xspace}
\newcommand{\sleec}{\textsc{sleec}\xspace}
\newcommand{\fastxml}{\textsc{fastxml}\xspace}
\newcommand{\fasttext}{\textsc{fasttext}\xspace}
\newcommand{\bert}{\textsc{bert}\xspace}
\newcommand{\elmo}{\textsc{elmo}\xspace}
\newcommand{\ulmfit}{\textsc{ulmfit}\xspace}

\newcommand{\nlp}{\textsc{nlp}\xspace}
\newcommand{\mimicii}{\textsc{mimic-ii}\xspace}
\newcommand{\mimiciii}{\textsc{mimic-iii}\xspace}
\newcommand{\tfidf}{\textsc{tf-idf}\xspace}

\newcommand{\encoder}{\textsc{enc}}
\newcommand{\decoder}{\textsc{dec}}

\pgfplotsset{compat=1.13}

\aclfinalcopy 

\title{Extreme Multi-Label Legal Text Classification:\\A case study in EU Legislation}

\author{Ilias Chalkidis* \qquad Manos Fergadiotis* \qquad Prodromos Malakasiotis* \\ \textbf{Nikolaos Aletras**} \qquad \textbf{Ion Androutsopoulos*} \\ * Department of Informatics, Athens University of Economics and Business, Greece \\
** Computer Science Department, University of Sheffield, UK\\ 
{\tt {\normalsize[ihalk,fergadiotis,rulller,ion]@aueb.gr, n.aletras@sheffield.ac.uk}}}

\date{}

\begin{document}
\maketitle
\begin{abstract}
We consider the task of Extreme Multi-Label Text Classification (\xmtc) in the legal domain. We release a new dataset of 57k legislative documents from \eurlex, the European Union's public document database, annotated with concepts from \eurovoc, a multidisciplinary thesaurus. The dataset is substantially larger than previous \eurlex datasets and suitable for \xmtc, few-shot and zero-shot learning. Experimenting with several neural classifiers, we show that \bigru{s} with self-attention outperform the current multi-label state-of-the-art methods, which employ label-wise attention. Replacing \cnn{s} with \bigru{s} in label-wise attention networks leads to the best overall performance.
\end{abstract}

\section{Introduction}
Extreme multi-label text classification (\xmtc), is the task of tagging documents with relevant labels from an extremely large label set, typically containing thousands of labels (classes). Applications include building web directories \cite{Partalas2015LSHTCAB}, labeling scientific publications with concepts from ontologies \cite{Tsatsaronis2015}, product categorization \cite{McAuley2013}, categorizing medical examinations \cite{Mullenbach2018,Rios2018-2}, and indexing legal documents \cite{Mencia2007}. We focus on legal text processing, an emerging \textsc{nlp} field with many applications \cite{Nallapati2008,Aletras2016,Chalkidis2017}, but limited publicly available resources.

We release a new dataset, named \newdata, including 57,000 English documents of \eu legislation from the \eurlex portal. All documents have been tagged with concepts from the European Vocabulary (\eurovoc), maintained by the Publications Office of the European Union. Although \eurovoc contains more than 7,000 concepts, most of them are rarely used in practice. Consequently, they are under-represented in \newdata, making the dataset also appropriate for few-shot and zero-shot learning.

Experimenting on \newdata, we explore the use of various \rnn-based and \cnn-based neural classifiers, including the state of the art Label-Wise Attention Network of \citet{Mullenbach2018}, called \lwancnn here. We show that both a simpler \textsc{bigru} with self-attention \cite{Xu2015} and the Hierarchical Attention Network (\han) of \citet{Yang2016} outperform \lwancnn by a wide margin. 
Replacing the \cnn encoder of \lwancnn with a \bigru, which leads to a method we call \lwangru, further improves performance. Similar findings are observed in the zero-shot setting where \zlwangru outperforms \zlwancnn.

\section{Related Work}
\label{sec:relatedwork}

\citet{Liu2017} proposed a \cnn similar to that of \citet{Kim2014} for \xmtc. They reported results 
on several benchmark datasets, most notably: \rcv \cite{Lewis2004}, containing news articles;
\eurlex \cite{Mencia2007}, containing legal documents; Amazon-12K \cite{McAuley2013}, containing
product descriptions; and Wiki-30K \cite{Zubiaga2012}, containing Wikipedia articles. Their proposed method outperformed both tree-based methods (e.g., \fastxml, \cite{Prabhu2014}) and target-embedding methods (e.g., \sleec \cite{Bhatia2015}, \fasttext \cite{bojanowski2016}).

\rnn{s} with self-attention have been employed in a wide variety of \nlp tasks, such as Natural Language Inference \cite{Liu2016}, Textual Entailment \cite{Rocktaschel2015}, and Text Classification \cite{Zhou2016}. \citet{You2018} used \rnn{s} with self-attention in \xmtc comparing with tree-based methods and deep learning approaches including vanilla \lstm{s} and \cnn{s}. Their method outperformed the other approaches in three out of four \xmtc datasets, demonstrating the effectiveness of attention-based \rnn{s}. 

\citet{Mullenbach2018} investigated the use of label-wise attention mechanisms in medical code prediction on the \mimicii and \mimiciii datasets \cite{Johnson2017}. \mimicii and \mimiciii contain over 20,000 and 47,000 documents tagged with approximately 9,000 and 5,000 \icd code descriptors, respectively. Their best method, Convolutional Attention for Multi-Label Classification, called \lwancnn here, includes multiple attention mechanisms, one for each one of the $L$ labels. 
\lwancnn outperformed weak baselines, namely logistic regression, vanilla \bigru{s} and \cnn{s}. Another important fact is that \lwancnn was found to have the best interpretability in comparison with the rest of the methods in human readers' evaluation.

\citet{Rios2018-2} discuss the challenge of few-shot and zero-shot learning on the \textsc{mimic} datasets. Over 50\% of all \icd labels never appear in \mimiciii, while 5,000 labels occur fewer than 10 times. The same authors proposed a new method, named Zero-Shot Attentive \cnn, called \zlwancnn here, which is similar to \lwancnn \cite{Mullenbach2018}, but also exploits the provided \icd code descriptors. The proposed \zlwancnn method was compared with prior state-of-the-art methods, including \lwancnn \cite{Mullenbach2018} and \matchcnn \cite{Rios2018-1}, a multi-head matching \cnn. While \zlwancnn did not outperform \lwancnn overall on \mimicii and \mimiciii, it had exceptional results in few-shot and zero-shot learning, being able to identify labels with few or no instances at all in the training sets. Experimental results showed an improvement of approximately four orders of magnitude in comparison with \lwancnn in few-shot learning and an impressive 0.269 $R@5$ in zero-shot learning, compared to zero $R@5$ reported for the other models compared.\footnote{See Section \ref{sec:measures} for a definition of $R@K$.} \citet{Rios2018-2} also apply graph convolutions to hierarchical relations of the labels, which improves the performance on few-shot and zero-shot learning. In this work, we do not consider relations between labels and do not discuss this method further.

Note that \lwancnn and \zlwancnn were not compared so far with strong generic text classification baselines. Both \citet{Mullenbach2018} and \citet{Rios2018-2} proposed sophisticated attention-based architectures, which intuitively are a good fit for \xmtc, but they did not directly compare those models with \rnn{s} with self-attention \cite{You2018} or even more complex architectures, such as Hierarchical Attention Networks (\han{s}) \cite{Yang2016}.

\section{EUROVOC \& EURLEX57K}

\subsection{EUROVOC Thesaurus}

\eurovoc is a multilingual thesaurus maintained by the Publications Office of the European Union.\footnote{\url{https://publications.europa.eu/en/web/eu-vocabularies}} It is used by the European Parliament, the national and regional parliaments in Europe, some national government departments, and other European organisations. The current version of \eurovoc contains more than 7,000 concepts referring to various activities of the \eu and its Member States (e.g., economics, health-care, trade, etc.). It has also been used for indexing documents in systems of \eu institutions, e.g., in web legislative databases, such as \eurlex and \cellar. All \eurovoc concepts are represented as tuples called \emph{descriptors}, each containing a unique numeric identifier and a (possibly) multi-word description of the concept concept, for example (1309, import), (693, citrus fruit), (192, health control), (863, Spain), (2511, agri-monetary policy).

\subsection{EURLEX57K}
\label{sec:dataset}

\newdata can be viewed as an improved version of the \eurlex dataset released by \citet{Mencia2007}, 
which included 19,601 documents tagged with 3,993 different \eurovoc concepts.  While \eurlex has been widely used in \xmtc research, it is less than half the size of \newdata and one of the smallest among \xmtc benchmarks.\footnote{The most notable \xmtc benchmarks can be found at \url{http://manikvarma.org/downloads/XC/XMLRepository.html}.} Over the past years the \eurlex archive has been widely expanded. \newdata is a more up to date dataset including 57,000 pieces of \eu legislation from the \eurlex portal.\footnote{\url{https://eur-lex.europa.eu}} All documents have been annotated by the Publications Office of \eu with multiple concepts from the \eurovoc 
thesaurus. \newdata is split in training (45,000 documents), development (6,000), and validation (6,000) subsets (see Table~\ref{tab:dataset}).\footnote{Our dataset is available at \url{http://nlp.cs.aueb.gr/software_and_datasets/EURLEX57K}, with permission of reuse under European Union\copyright, \url{https://eur-lex.europa.eu}, 1998--2019.}

\begin{table}[ht]
\centering
\footnotesize
\begin{tabular}{lcccc}
  Subset & Documents ($D$) & Words/$D$ & Labels/$D$ \\
\hline
  Train & 45,000 & 729  & 5\\
  Dev. & 6,000 & 714  & 5 \\
  Test & 6,000 & 725  & 5\\
  \hline
\end{tabular}
\caption{Statistics of the \textsc{\eurlex} dataset.}
\vspace*{-4mm}
\label{tab:dataset}
\end{table}

All documents are structured in four major zones: the \emph{header} including the title and the name of the legal body that enforced the legal act; the \emph{recitals} that consist of references in the legal background of the decision; the \emph{main body}, which is usually organized in articles; and the \emph{attachments} that usually include appendices and annexes. For simplicity, we will refer to each one of \emph{header}, \emph{recitals}, \emph{attachments} and each of the \emph{main body}'s articles as \emph{sections}. We have pre-processed all documents in order to provide the aforementioned structure. 

\begin{figure}
  \centering
    \includegraphics[width=\columnwidth]{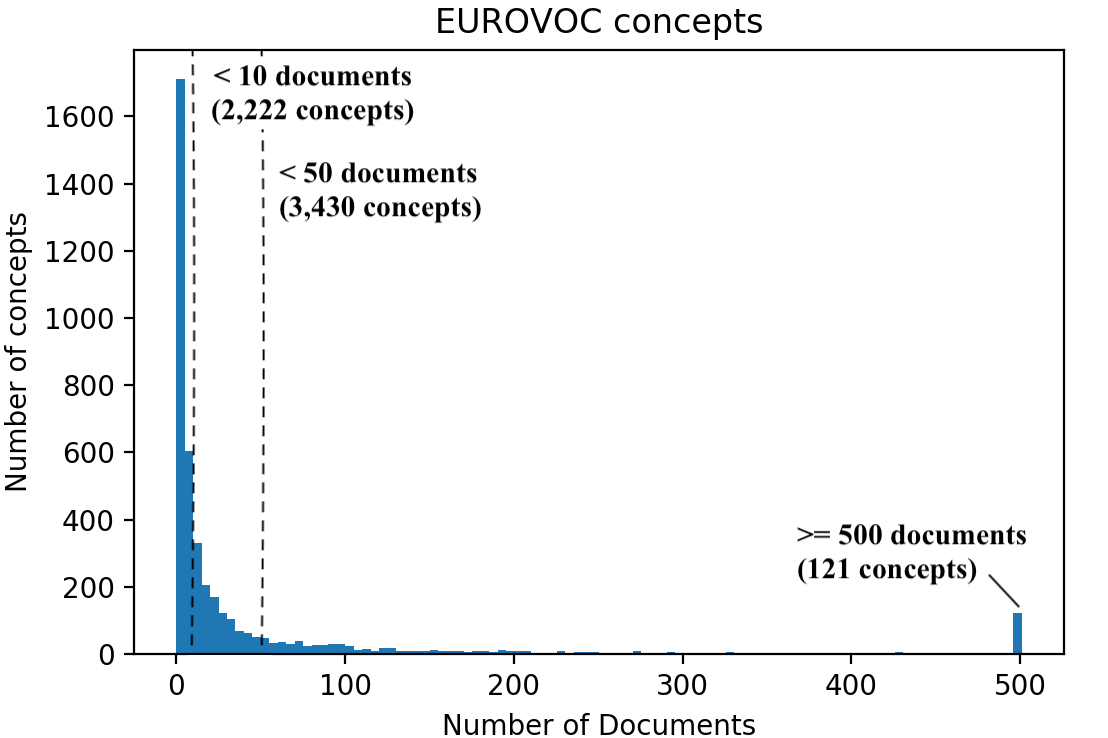}
  \caption{EUROVOC concepts frequency.}
  \label{fig:histogram}
\end{figure}

While \eurovoc includes over 7,000 concepts (labels), only 4,271 (59.31\%) of them are present in \newdata. Another important fact is that most labels are under-represented; only 2,049 (47,97\%) have been assigned to more than 10 documents. Such an aggressive Zipfian distribution (Figure~\ref{fig:histogram}) has also been noted in other domains, like medical examinations \cite{Rios2018-2} where \xmtc has been applied to index documents with concepts from medical thesauri. 

The labels of \newdata are divided in three categories: \emph{frequent} labels (746), which occur in more than 50 training documents and can be found in all three subsets (training, development, test); 
\emph{few-shot} labels (3,362), which appear in 1 to 50 training documents; and \emph{zero-shot} labels (163), which appear in the development and/or test, but not in the training, documents.

\begin{figure*}[ht]
  \centering
    \includegraphics[width=\textwidth]{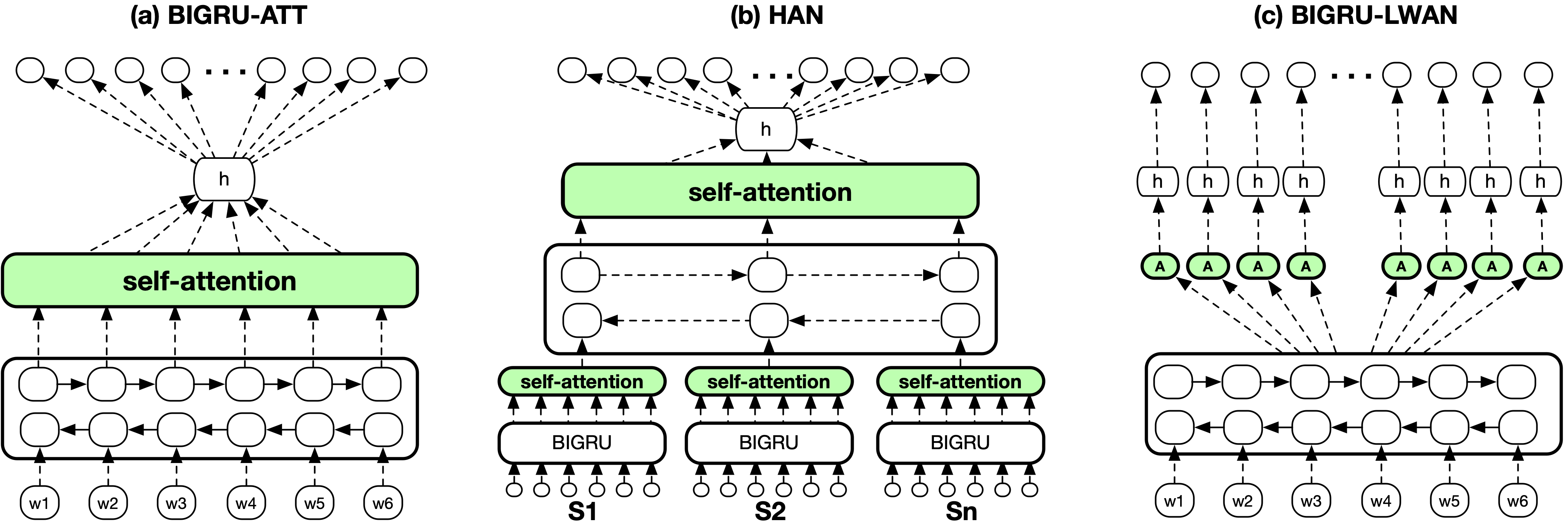}
  \caption{Illustration of (a) \bigruatt, (b) \han, and (c) \lwangru.}
  \label{fig:methods}
\end{figure*}

\section{Methods Considered}
We experiment with a wide repertoire of methods including linear and non-linear neural classifiers. We also propose and conduct initial experiments with two novel neural methods that aim to cope with the extended length of the legal documents and the information sparsity (for \xmtc purposes) across the \emph{sections} of the documents.

\subsection{Baselines}

\subsubsection{Exact Match}
To demonstrate that plain label name matching is not sufficient, our first weak baseline, Exact Match, tags documents only with labels whose descriptors appear verbatim in the documents.

\subsubsection{Logistic Regression}
To demonstrate the limitations of linear classifiers with bag-of-words representations, we train a Logistic Regression classifier with \tfidf scores for the most frequent unigrams, bigrams, trigrams, 4-grams, 5-grams across all documents. Logistic regression with similar features has been widely used for multi-label classification in the past.

\subsection{Neural Approaches}

We present eight alternative neural methods. In the following subsections, we describe their structure consisting of five main parts: 
\begin{itemize}
	\item \emph{word encoder} (\encoder$_w$): turns word embeddings into context-aware embeddings,
    \item \emph{section encoder} (\encoder$_s$): turns each section (sentence) into a sentence embedding,
    \item \emph{document encoder} (\encoder$_d$): turns an entire document into a final dense representation,
    \item \emph{section decoder} (\decoder$_s$) or \emph{document decoder} (\decoder$_d$): maps the section or document representation to a many-hot label assignment.
\end{itemize}
All parts except for \encoder$_w$ and \decoder$_d$ are optional, i.e., they may not be present in all methods.

\subsubsection{BIGRU-ATT}
In the first deep learning method, \bigruatt (Figure~\ref{fig:methods}a), \encoder$_w$ is a stack of \bigru{s} that converts the pre-trained word embeddings ($w_{t}$) to context-aware ones ($h_{t}$). \encoder$_d$ employs a self attention mechanism to produce the final representation $d$ of the document as a weighted sum of $h_{t}$: 

\vspace{-1mm}
\begin{eqnarray}
a_t &=& \frac{\mathrm{exp}(h_t^\top u)}{\sum_j\mathrm{exp}(h_j^\top u)}\\
d &=& \frac{1}{T} \sum^T_{t=1} a_t h_{t}
\label{att_eq}
\end{eqnarray}
$T$ is the document's length in words, and $u$ is a trainable vector used to compute the attention scores $a_{t}$ over $h_{t}$. \decoder$_d$ is a linear layer with $L=4,271$ output units and 
sigmoid ($\sigmoid$) activations that maps the document representation $d$ to $L$ probabilities, one per label.

\subsubsection{HAN}

The Hierarchical Attention Network (\han) \cite{Yang2016}, exploits the structure of the documents by encoding the text in two consecutive steps (Figure~\ref{fig:methods}b). First, a \bigru (\encoder$_w$) followed by a self-attention mechanism (\encoder$_s$) turns the word embeddings ($w_{it}$) of each section $s_i$ with $T_i$ words into a section embedding $c_i$: 

\vspace{-1mm}
\begin{eqnarray}
v_{it} &=& \tanh(
W^{(s)} h_{it} + b^{(s)})\\
a^{(s)}_{it} &=& \frac{\mathrm{exp}(v_{it}^\top
u^{(s)})}{\sum_j\mathrm{exp}(v_{ij}^\top 
u^{(s)})}\\
c_i &=& 
\frac{1}{T_i} \sum^{T_i}_{t=1} a^{(s)}_{it} h_{it}
\end{eqnarray}
where $u^{(s)}$ is a trainable  vector. Next, \encoder$_d$, another \bigru with self-attention, converts the section embeddings ($S$ in total, as many as the sections) to the final document representation $d$:
\vspace{-1mm}
\begin{eqnarray}
v_i &=& \tanh(W^{(d)} c_i + b^{(d)})\\
a^{(d)}_{i} &=& \frac{\mathrm{exp}(v_{i}^\top u^{(d)})}{\sum_j\mathrm{exp}(v_{j}^\top u^{(d)})}\\
d &=& \frac{1}{S} \sum^S_{i=1} a^{(d)}_{i} c_{i}
\end{eqnarray}

\noindent where $u^{(d)}$ is a trainable vector. The final decoder \decoder$_d$ of \han is the same as in \bigruatt.

\subsection{MAX-HSS}
\label{max-hss}
Initial experiments we conducted indicated that \han is  outperformed by the shallower \bigruatt. We suspected that the main reason was the fact that the section embeddings $c_i$ that \han's \encoder$_s$ produces contain useful information that is later degraded by \han's \encoder$_d$. Based on this assumption, we experimented with a novel method, named Max-Pooling over Hierarchical Attention Scorers (\maxhss). \maxhss produces section embeddings $c_i$ in the same way as \han, but then employs a separate \decoder$_s$ per section to produce label predictions from each section embedding $c_i$:
\vspace{-1mm}
\begin{equation}
p^{(s)}_i = \sigmoid (W^{(m)} c_i + b^{(m)})
\end{equation}
where $p_i$ is an $L$-dimensional vector containing probabilities for all labels, derived from $c_i$. \decoder$_d$ aggregates the predictions for the whole document with a $\maxpool$  
operator that extracts the highest probability per label across all sections:  
\vspace{-1mm}
\begin{equation}
p^{(d)} = \maxpool(
p^{(s)}_1, \dots, p^{(s)}_S)
\end{equation}
Intuitively, each section tries to predict the labels relying on its content independently, and \decoder$_d$ extracts the most probable labels across sections.

\subsubsection{CNN-LWAN and BIGRU-LWAN}
The Label-wise Attention Network, \lwan \cite{Mullenbach2018}, also uses a self-attention mechanism, but here \encoder$_d$ employs $L$ independent attention heads, one per label, generating $L$ document representations $d_l = \sum_t a_{lt} h_t$ ($l=1, \dots, L$) from the sequence of context aware word embeddings $h_1, \dots, h_T$ of each document $d$.
The intuition is that each attention head focuses on possibly different aspects of $h_1, \dots, h_T$ needed to decide if the corresponding label should be assigned to the document or not. \decoder$_d$ employs $L$ linear layers with $\sigmoid$ activation, each one operating on a label-wise document representation $d_l$ to produce the probability for the corresponding label. In the original \lwan \citep{Mullenbach2018}, called \lwancnn here, \encoder$_w$ is a vanilla \cnn. We use a modified version, \lwangru, where \encoder$_w $ is a \bigru (Figure~\ref{fig:methods}c).

\subsection{Z-CNN-LWAN and Z-BIGRU-LWAN}
\label{sec:zero_shot_methods}

Following the work of \citet{Mullenbach2018}, \citet{Rios2018-2} designed a similar architecture in order to improve the results in documents that are classified with rare labels. In one of their models, \encoder$_d$ creates label representations, $u_l$, from the corresponding descriptors as follows:
\begin{equation} 
u_l  = \frac{1}{E} \sum^E_{e=1} w_{le}
\label{eq:labelEmbeddings}
\end{equation}
where $w_{le}$ is the word embedding of the $e$-th word in the $l$-th label descriptor. The label representations are then used as alternative attention vectors:

\vspace{-2mm}
\begin{eqnarray}
v_t &=& \tanh(W^{(z)} h_t + b^{(z)}) \\
a_{lt} &=& \frac{\mathrm{exp}(v_t^\top u_{l})}{
\sum_j \mathrm{exp}(v_j^\top u_{l})}
\label{eq:alpha_lt}\\
d_l &=&  \frac{1}{T} \sum^T_{t=1} a_{lt} h_t
\end{eqnarray}
where $h_t$ are the context-aware embeddings produced by a vanilla \cnn (\encoder$_w$) operating 
on the document's word embeddings, $a_{lt}$ are the attention scores conditioned on the corresponding label representation $u_l$, and $d_l$ is the label-wise document representation. \decoder$_d$ also relies on label representations to produce each label's probability:
\vspace{-2mm}
\begin{eqnarray}
p_l &=&  \sigmoid(u_l^\top d_l) 
\label{eq:p_l}
\end{eqnarray}

Note that the representations $u_l$ of both encountered (during training) and unseen (zero-shot) labels remain unchanged, because the word embeddings $w_{le}$ are not updated (Eq.~\ref{eq:labelEmbeddings}). This keeps the representations of zero-shot labels close to those of encountered labels they share several descriptor words with. In turn, this helps the attention mechanism (Eq.~\ref{eq:alpha_lt}) and the decoder (Eq.~\ref{eq:p_l}), where the label representations $u_l$ are used, cope with unseen labels that have similar descriptors with encountered labels. As with \lwancnn and \lwangru, we experiment with the original version of the model of \citet{Rios2018-2}, which uses a \cnn \encoder$_w$  (\zlwancnn), and a version that uses a \bigru \encoder$_w$ (\zlwangru).

\subsection{LW-HAN}

We also propose a new method, Label-Wise Hierarchical Attention Network (\lwhan), that combines ideas from both \han and \lwan. For each section, \lwhan employs an \lwan to produce $L$ probabilities. Then, like \maxhss, a $\maxpool$ operator extracts the highest probability per label across all sections. In effect, \lwhan exploits the document structure to cope with the extended document length of legal documents, while employing multiple label-wise attention heads to deal with the vast and sparse label set. By contrast, \maxhss does not use label-wise attention.

\section{Experimental Results}
\label{experiments}

\subsection{Experimental Setup}

 Hyper-parameters were tuned on development data using \textsc{hyperopt}.\footnote{\url{https://github.com/hyperopt}} We tuned for the following hyper-parameters and ranges: \encoder\xspace output units \{200, 300, 400\}, \encoder\xspace layers \{1, 2\}, batch size \{8, 12, 16\}, dropout rate \{0.1, 0.2, 0.3, 0.4\}, word dropout rate \{0.0, 0.01, 0.02\}. For the best hyper-parameter values, we perform five runs and report mean scores on test data. For statistical significance, we take the run of each method with the best performance on development data, and perform two-tailed approximate randomization tests \cite{Dror2018} on test data. We used 200-dimensional pre-trained \glove embeddings \cite{pennington2014glove} in all neural methods.

\subsection{Evaluation Measures} 
\label{sec:measures}
The most common evaluation measures in \xmtc are recall ($R@K$), precision ($P@K$), and \ndcg
($\ndcg@K$) at the top $K$ predicted labels, along with micro-averaged $F$-1 across all labels. 
Measures that macro-average over labels do not consider the number of instances per label,
thus being very sensitive to infrequent labels, which are many more than frequent ones (Section~\ref{sec:dataset}). On the other hand, ranking measures, like $R@K$, $P@K$, $\ndcg@K$, 
are sensitive to the choice of $K$. In \newdata the average number of labels per document is 5.07,
hence evaluating at $K=5$ is a reasonable choice. We note that 99.4\% of the dataset's documents have at most 10 gold labels. 

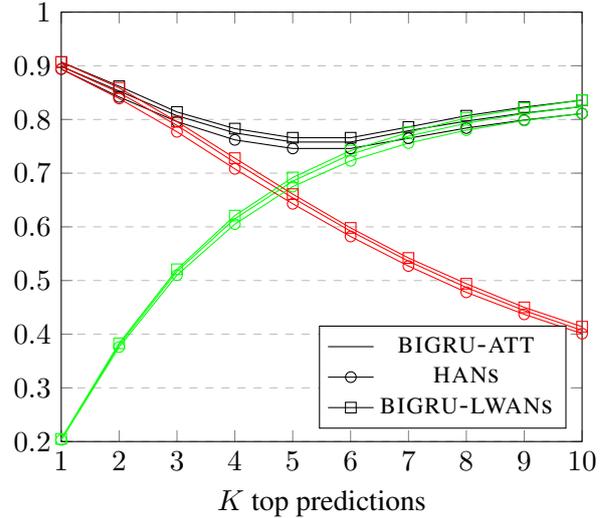
\begin{figure}[ht]
\centering
\begin{tikzpicture}[scale=1.0]
    \begin{axis}[
        xlabel={$K$ top predictions},
        ylabel={},
        xmin=1, xmax=10,
        ymin=0.2, ymax=1,
        xtick={1,2,3,4,5,6,7,8,9,10},
        ytick={0.2, 0.3, 0.4, 0.5, 0.6, 0.7, 0.8, 0.9, 1.0},
        legend pos=south east,
        ymajorgrids=true,
        grid style=dashed,
    ]
    
    \addplot[
            color=black,
            ]
            coordinates {
            (1,0.899)(2,0.853)(3,0.807)(4,0.775)(5,0.758)(6,0.758)(7,0.779)(8,0.797)(9,0.812)(10,0.824)};
            \addlegendentry{\bigruatt}
            
    \addplot[
            color=black,
            mark=o,
            ]
            coordinates {
            (1,0.894)(2,0.842)(3,0.796)(4,0.762)(5,0.746)(6,0.746)(7,0.765)(8,0.784)(9,0.799)(10,0.811)};
            \addlegendentry{\han{s}}
            
    \addplot[
            color=black,
            mark=square,
            ]
            coordinates {
            (1,0.907)(2,0.862)(3,0.814)(4,0.783)(5,0.766)(6,0.766)(7,0.786)(8,0.807)(9,0.823)(10,0.836)};
            \addlegendentry{\lwangru{s}}
     
    \addplot[
        color=red,
                ]
        coordinates {
        (1,0.899)(2,0.850)(3,0.789)(4,0.720)(5,0.654)(6,0.592)(7,0.536)(8,0.487)(9,0.444)(10,0.407)};

    \addplot[
        color=green,
                ]
        coordinates {
        (1,0.204)(2,0.380)(3,0.517)(4,0.615)(5,0.685)(6,0.735)(7,0.769)(8,0.793)(9,0.811)(10,0.824)};
    
    \addplot[
        color=red,
        mark=o,
        ]
        coordinates {
        (1,0.894)(2,0.839)(3,0.777)(4,0.708)(5,0.643)(6,0.582)(7,0.527)(8,0.478)(9,0.437)(10,0.401)};

    \addplot[
        color=green,
        mark=o,
        ]
        coordinates {
        (1,0.203)(2,0.376)(3,0.510)(4,0.605)(5,0.675)(6,0.723)(7,0.756)(8,0.780)(9,0.798)(10,0.811)};
    
    \addplot[
        color=red,
        mark=square,
        ]
        coordinates {
        (1,0.907)(2,0.859)(3,0.796)(4,0.728)(5,0.661)(6,0.598)(7,0.542)(8,0.494)(9,0.450)(10,0.414)};

    \addplot[
        color=green,
        mark=square,
        ]
        coordinates {
        (1,0.205)(2,0.383)(3,0.521)(4,0.621)(5,0.692)(6,0.742)(7,0.776)(8,0.803)(9,0.821)(10,0.836)};
     
\end{axis}
\end{tikzpicture}
\caption{$R@K$ (green lines), $P@K$ (red), $RP@K$ (black) scores of the best methods (\bigruatt, \han{s}, \lwangru), for $K=1$ to 10. All scores macro-averaged over test documents.}
\label{fig:kappa_comparison}
\end{figure}

While $R@K$ and $P@K$ are commonly used, we question their suitability for \xmtc. $R@K$ leads to unfair penalization of methods when documents have more than $K$ gold labels. Evaluating at $K=1$ for a document with $N > 1$ gold labels returns at most $R@1=\frac{1}{N}$, unfairly penalizing systems by not allowing them to return $N$ labels. This is shown in Figure~\ref{fig:kappa_comparison}, where the green lines show that $R@K$ decreases as $K$ 
decreases, because of low scores obtained for documents with more than $K$ labels. On the other hand, $P@K$ leads to excessive penalization for documents with fewer than $K$ gold labels. Evaluating at $K=5$ for a document with just one gold label returns at most $P@5=\frac{1}{5} =0.20$, 
unfairly penalizing systems that retrieved all the gold labels (in this case, just one). The 
red lines of Figure~\ref{fig:kappa_comparison} decline as $K$ increases, because the number of documents with fewer than $K$ gold labels increases (recall that the average number of gold labels is 5.07).

Similar concerns have led to the introduction of $\mathrm{R}\text{-}\mathrm{Precision}$ and $\ndcg@K$ in Information Retrieval \cite{Manning2009}, which we believe are also more appropriate for \xmtc. Note, however, that $\mathrm{R}\text{-}\mathrm{Precision}$ requires that the number of gold labels per document is known beforehand, which is not realistic in practical applications. Therefore we propose $\mathrm{R}\text{-}\mathrm{Precision}@K$ ($RP@K$) where $K$ is the maximum number of retrieved labels. Both $RP@K$ and $\ndcg@K$ adjust to the number of gold labels per document, without unfairly penalizing systems for documents with fewer than $K$  or many more than $K$ gold labels. They are defined as follows:
\vspace{-0.3em}
\begin{eqnarray}
RP@K = \frac{1}{N} \sum^N_{n=1}
\sum^K_{k=1}\frac{\mathrm{Rel}(n, k)}{\min{(K,R_n)}}\\
\ndcg@K = \frac{1}{N} \sum^N_{n=1} Z_{Kn} \sum^K_{k=1} \frac{2^{\mathrm{Rel}(n, k)}-1}{
\log_2{(1+k)}} 
\end{eqnarray}
Here $N$ is the number of test documents; $\mathrm{Rel}(n ,k)$ is 1 if the $k$-th retrieved label of the $n$-th test document is correct, otherwise 0; $R_n$ is the number of gold labels of the $n$-th test document; and $Z_{Kn}$ is a normalization factor to ensure that $\ndcg@K = 1$ for perfect ranking.

In effect, $RP@K$ is a macro-averaged (over test documents) version of $P@K$, but $K$ is reduced to the number of gold labels $R_n$ of each test document, if $K$ exceeds $R_n$. Figure~\ref{fig:kappa_comparison} shows $RP@K$ for the three best systems. Unlike $P@K$, $RP@K$ does not decline sharply as $K$ increases, because it replaces $K$ by $R_n$ (number of gold labels) when $K>R_n$. For $K=1$, $RP@K$ is equivalent to $P@K$, as confirmed by Fig.~\ref{fig:kappa_comparison}. For large values of $K$ that almost always exceed $R_n$, $RP@K$ asymptotically approaches $R@K$ (macro-averaged over documents), as also confirmed by Fig.~\ref{fig:kappa_comparison}.

\begin{table*}[ht!]
\centering
{
\footnotesize\addtolength{\tabcolsep}{-2pt}
\begin{tabular}{lccccccccc}
  \hline
  & \multicolumn{3}{c}{\textsc{All Labels}} & \multicolumn{2}{c}{\textsc{Frequent}} & \multicolumn{2}{c}{\textsc{Few}} & \multicolumn{2}{c}{\textsc{Zero}} \\ 
  & $RP@5$ & $nDCG@5$ & Micro-$F1$ & $RP@5$ & $nDCG@5$ & $RP@5$ & $nDCG@5$ & $RP@5$ & $nDCG@5$ \\
  \cline{2-10}
  Exact Match & 0.097 & 0.099 & 0.120 & 0.219 & 0.201 & 0.111 & 0.074 & 0.194 & 0.186 \\
  Logistic Regression & 0.710 & 0.741 & 0.539 & 0.767 & 0.781 & 0.508 & 0.470 & 0.011 & 0.011 \\
  \hline
  \bigruatt & 0.758 & 0.789 & 0.689 & 0.799 & 0.813 & 0.631 & 0.580 & 0.040 & 0.027\\
  \han & 0.746 & 0.778 & 0.680 & 0.789 & 0.805 & 0.597 & 0.544 & 0.051 & 0.034\\
  \hline
 \lwancnn & 0.716 & 0.746 & 0.642 & 0.761 & 0.772 & 0.613 & 0.557 & 0.036  & 0.023 \\
  \lwangru & \textbf{0.766} & \textbf{0.796} & \textbf{0.698} & \textbf{0.805} & \textbf{0.819} & \textbf{0.662} & \textbf{0.618} & 0.029 & 0.019\\
   \hline
  \zlwancnn & 0.684 & 0.717 & 0.618 & 0.730 & 0.745 & 0.495 & 0.454 & 0.321 & 0.264 \\
  \zlwangru & 0.718 & 0.752 & 0.652  & 0.764 & 0.780 & 0.561 & 0.510 & \textbf{0.438} & \textbf{0.345} \\
  \hline
   \ensemble & \textbf{0.766} & \textbf{0.796} & \textbf{0.698} & \textbf{0.805} & \textbf{0.819} & \textbf{0.662} & \textbf{0.618} & \textbf{0.438} & \textbf{0.345} \\
  \hline
  \maxhss & 0.737 & 0.773 & 0.671 & 0.784 & 0.803 & 0.463 & 0.443 & 0.039 & 0.028 \\
  \lwhan & 0.721 & 0.761 & 0.669 & 0.766 & 0.790 & 0.412 & 0.402 & 0.039 & 0.026 \\
  \hline
\end{tabular}
}
\caption{Results on \newdata for all, frequent ($>50$ training instances), few-shot (1 to 50 instances), and zero-shot labels.  All the differences between the best (bold) and other methods are statistically significant ($p < 0.01$).}
\vspace*{-4mm}
\label{tab:results}
\end{table*}

\subsection{Overall Experimental Results}
\label{sec:overall}

Table~\ref{tab:results} reports experimental results for all methods and evaluation measures. As expected, Exact Match is vastly outperformed by machine learning methods, while Logistic Regression is also unable to cope with the complexity of \xmtc.

In Section~\ref{sec:relatedwork}, we referred to the lack of previous experimental comparison between methods relying on label-wise attention and strong generic text classification baselines. Interestingly, for all, frequent, and even few-shot labels, the generic \bigruatt performs better than \lwancnn, which was designed for \xmtc. \han also performs better than \lwancnn for all and frequent labels. However, replacing the \cnn encoder of \lwancnn with a \bigru (\lwangru) leads to the best results overall, with the exception of zero-shot labels, indicating that the main weakness of \lwancnn is its vanilla \cnn encoder.

\subsection{Few-shot and Zero-shot Results}

\begin{figure*}[ht!]
\centering
\begin{subfigure}[b]{1\textwidth}
\includegraphics[width=1\linewidth]{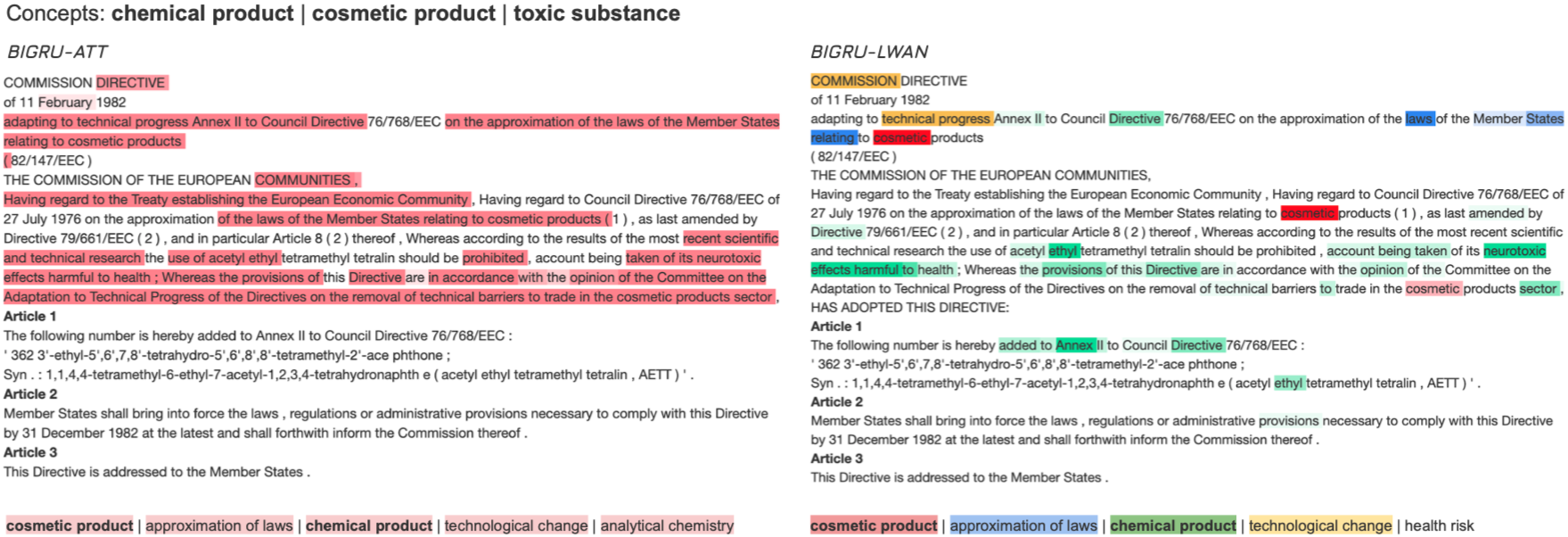}
\caption{COMMISSION DIRECTIVE (EEC) No 82/147}
\label{fig:attention}
\end{subfigure}
\begin{subfigure}[b]{1\textwidth}
\centering
\includegraphics[width=1\linewidth]{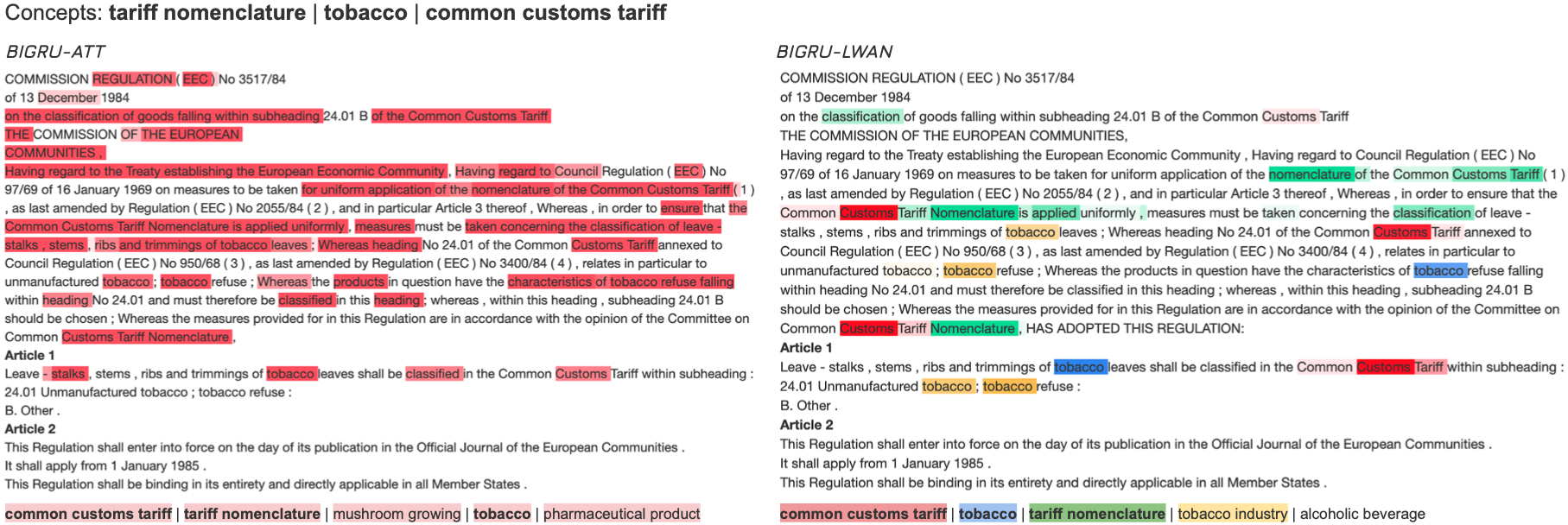}
\caption{COMMISSION REGULATION (EEC) No 3517/84}
\label{fig:label_wise_attention}
\end{subfigure}
\caption{Attention heat-maps for \bigruatt (left) and \lwangru (right). 
Gold labels (concepts) are shown at the top of each sub-figure, while the top 5 predicted labels are shown at the bottom. Correct predictions are shown in bold. \lwangru's label-wise attentions are depicted in different colors.}
\end{figure*}

As noted by Rios and Kavuluru \shortcite{Rios2018-2}, developing reliable and robust classifiers for few-shot and zero-shot tasks is a significant challenge. Consider, for example, a test document referring to concepts that have rarely (few-shot) or never (zero-shot) occurred in training documents (e.g., `tropical disease', which exists once in the whole dataset). A reliable classifier should be able to at least make a good guess for such rare concepts.

As shown in Table~\ref{tab:results}, \lwangru outperforms all other methods in both frequent and few-shotlabels, but not in zero-shot labels, where \zlwancnn \cite{Rios2018-2} provides exceptional results compared to other methods. Again, replacing the vanilla \cnn of \zlwancnn with a \bigru (\zlwangru) improves performance across all label types and measures.  

All other methods, including \bigruatt, \han, \lwan, fail to predict relevant zero-shot labels (Table~\ref{tab:results}). This behavior is not surprising, because the training objective, minimizing binary cross-entropy across all labels, largely ignores infrequent labels. The zero-shot versions of \lwancnn and \lwangru outperform all other methods on zero-shot labels, in line with the findings of \citet{Rios2018-2}, because they exploit label descriptors, which they do not update during training (Section~\ref{sec:zero_shot_methods}). Exact Match also performs better than most other methods (excluding \zlwancnn and \zlwangru) on zero-shot labels, because it exploits label descriptors.

To better support all types of labels (frequent, few-shot, zero-shot), we propose an ensemble of
\lwangru and \zlwangru, which outputs the predictions of \lwangru for frequent and few-shot labels, along with the predictions of \zlwangru for zero-shot labels. The ensemble's results for `all labels' in  Table~\ref{tab:results} are the same as those of \lwangru, because zero-shot labels are very few (163) and rare in the test set.

The two methods (\maxhss, \lwhan) that aggregate (via $\maxpool$) predictions across sections under-perform in all types of labels, suggesting that combining predictions from individual sections is not a promising direction for \xmtc.

\subsection{Providing Evidence through Attention}

\citet{Chalkidis2018} noted that self-attention does not only lead to performance improvements in legal text classification, but might also provide useful evidence for the predictions (i.e., assisting in decision-making). On the left side of Figure~\ref{fig:attention}, we demonstrate such indicative results by visualizing the attention heat-maps of \bigruatt and \lwangru. Recall that \lwangru uses a separate attention head per label. This allows producing multi-color heat-maps (a different color per label) separately indicating which words the system attends most when predicting each label. By contrast,  \bigruatt uses a single attention head and, thus, the resulting heat-maps include only one color.

\section{Conclusions and Future Work}
We compared various neural methods on a new legal \xmtc dataset, \newdata, also investigating few-shot and zero-shot learning. We showed that \bigruatt is a strong baseline for this \xmtc dataset,
outperforming \lwancnn \cite{Mullenbach2018}, which was especially designed for \xmtc, but that replacing the vanilla \cnn of \lwancnn by a \bigru encoder (\lwangru) leads to the best overall results, except for zero-shot labels. For the latter, the zero-shot version of \lwancnn of \citet{Rios2018-2} produces exceptional results, compared to the other methods, and its performance improves further when its \cnn is replaced by a \bigru  (\zlwangru). Surprisingly \han \cite{Yang2016} and other hierarchical methods we considered (\maxhss, \lwhan) are weaker compared to the other neural methods we experimented with, which do not consider the structure (sections) of the documents. 

The best methods of this work rely on \gru{s} and thus are computationally expensive. The length of the documents further affects the training time of these methods. Hence, we plan to investigate the use of Transformers \cite{Vaswani2017, Dai2019} and dilated \cnn{s} \cite{KalchbrennerESO16}
as alternative document encoders.

Given the recent advances in transfer learning for natural language processing, we plan to experiment with pre-trained neural language models for feature extraction and fine-tuning using state-of-the-art approaches such as \elmo~\cite{Peters2018}), \ulmfit~\cite{ULMFit} and \bert~\cite{BERT}.

Finally, we also plan to investigate further the extent to which attention heat-maps provide useful explanations of the predictions made by legal predictive models following recent work on attention explainability \cite{Jain2019}.

\bibliography{naaclhlt2019}
\bibliographystyle{acl_natbib}

\end{document}